\documentclass[10pt,twocolumn,letterpaper]{article}

\usepackage{cvpr}              % To produce the CAMERA-READY version
\usepackage[accsupp]{axessibility}

\newcommand{\model}{EMGauss} % model name
\usepackage{multirow}
\usepackage{algorithm}
\usepackage{algpseudocode}
\usepackage{marvosym}

\definecolor{cvprblue}{rgb}{0.21,0.49,0.74}
\usepackage[pagebackref,breaklinks,colorlinks,allcolors=cvprblue]{hyperref}

\def\paperID{33421} % *** Enter the Paper ID here
\def\confName{CVPR}
\def\confYear{2026}

\title{\textit{EMGauss}: Continuous Slice-to-3D Reconstruction via Dynamic Gaussian Modeling in Volume Electron Microscopy}

\author{
Yumeng~He
\quad
Zanwei~Zhou
\quad
Yekun~Zheng
\quad
Chen~Liang
\quad
Yunbo~Wang\thanks{Corresponding author.}
\quad
Xiaokang~Yang
\\
MoE Key Lab of Artificial Intelligence, AI Institute, School of Computer Science
\\Shanghai Jiao Tong University\\
{\tt\small \{ymhe, sjtu19zzw, beta\_cat, no8irch6en, yunbow, xkyang\}@sjtu.edu.cn}
}

\begin{document}

%%%%%%%%%%%%%%%%%%%%%%%%%%%%%% main paper %%%%%%%%%%%%%%%%%%%%%%%%%%%%%%
\maketitle

\begin{abstract}
% We present SliceFlow, a general framework for 3D reconstruction from planar scanned 2D slices with applications in Volume Electron Microscopy (vEM). Our key innovation is to reframe slice-to-3D reconstruction as a video rendering task based on Gaussian splatting, modeling the progression of axial slices as the temporal evolution of 2D Gaussian point clouds. To capture spatial displacements, deformations, and density variations across slices, we leverage a PointTransformer with shared parameters, which promotes spatial coherence throughout the reconstructed volume (up to a $2k$ slice resolution). Compared to existing diffusion- and transformer-based methods for vEM reconstruction, SliceFlow significantly reduces test errors, supports continuous slice synthesis at arbitrary depths, and eliminates the need for large-scale pretraining, thereby reducing dependence on densely sampled vEM stacks. Beyond vEM, SliceFlow offers a scalable and generalizable solution for slice-to-3D reconstruction across diverse imaging domains.

Volume electron microscopy (vEM) enables nanoscale 3D imaging of biological structures but remains constrained by acquisition trade-offs, leading to anisotropic volumes with limited axial resolution. Existing deep learning methods seek to restore isotropy by leveraging lateral priors; yet their assumptions break down for morphologically anisotropic structures. We present \textbf{EMGauss}, a general framework for 3D reconstruction from planar scanned 2D slices with applications in vEM, which circumvents the inherent limitations of isotropy-based approaches. Our key innovation is to reframe slice-to-3D reconstruction as a 3D dynamic scene rendering problem based on Gaussian splatting, where the progression of axial slices is modeled as the temporal evolution of 2D Gaussian point clouds. To enhance fidelity in data-sparse regimes, we incorporate a \textbf{Teacher–Student bootstrapping mechanism} that uses high-confidence predictions on unobserved slices as pseudo-supervisory signals. Compared with diffusion- and GAN-based reconstruction methods, EMGauss substantially improves interpolation quality, enables continuous slice synthesis, and eliminates the need for large-scale pretraining. Beyond vEM, it potentially provides a generalizable slice-to-3D solution across diverse imaging domains.

\end{abstract}

% \begin{figure}[t]
%   \centering
%   \fbox{\rule{0pt}{2in} \rule{0.9\linewidth}{0pt}}
%    %\includegraphics[width=0.8\linewidth]{egfigure.eps}

%    \caption{Example of caption.
%    It is set in Roman so that mathematics (always set in Roman: $B \sin A = A \sin B$) may be included without an ugly clash.}
%    \label{fig:onecol}
% \end{figure}

% \begin{figure*}
%   \centering
%   \begin{subfigure}{0.68\linewidth}
%     \fbox{\rule{0pt}{2in} \rule{.9\linewidth}{0pt}}
%     \caption{An example of a subfigure.}
%     \label{fig:short-a}
%   \end{subfigure}
%   \hfill
%   \begin{subfigure}{0.28\linewidth}
%     \fbox{\rule{0pt}{2in} \rule{.9\linewidth}{0pt}}
%     \caption{Another example of a subfigure.}
%     \label{fig:short-b}
%   \end{subfigure}
%   \caption{Example of a short caption, which should be centered.}
%   \label{fig:short}
% \end{figure*}

\section{Introduction}
\label{sec:intro}

Volume electron microscopy (vEM) has revolutionized the understanding of biological ultrastructures, enabling nanoscale 3D imaging of cells, tissues, and even entire organisms~\cite{collinson2023volume,glausier2025volume,li2025situ,robert2025improving,turegano2024tracing}. 
However, due to the inherent trade-off among resolution, field of view, and acquisition time---commonly referred to as the ``impossible triangle''---direct acquisition of isotropic data remains costly and inefficient~\cite{heinrich2017deep,peddie2022volume}. This often results in anisotropic volumes, with axial (z) resolution substantially lower than the in-plane (xy) resolution.

\begin{figure}[t]
\centering
\includegraphics[width=\linewidth]{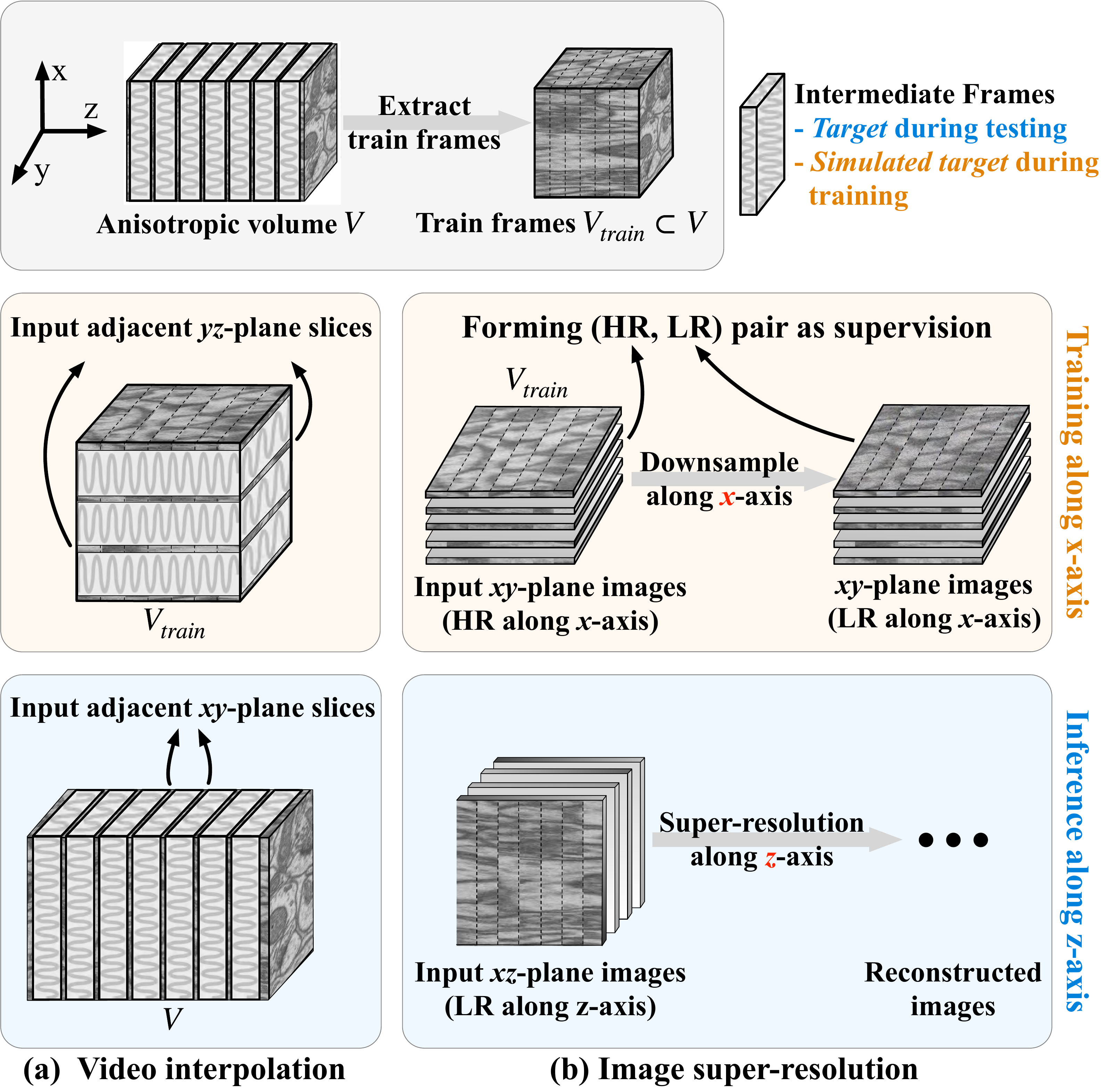}
\vspace{-10pt}
\caption{\textit{Different paradigms of isotropic reconstruction from anisotropic datasets.}
Existing approaches typically extract volume features from abundant xy-plane images. After stacking anisotropic slices along the z-axis (top), they:
(a) \textbf{Video interpolation along the z-axis}: models are trained along the \textit{x} or \textit{y} direction, where adjacent frames are input to synthesize the intermediate frame; during inference, the trained model is repurposed to interpolate along the \textit{z}-axis, generating intermediate slices corresponding to continuous temporal coordinates ($\tau\in[0,1]$).
(b) \textbf{Image super-resolution on xz/yz planes}: Since the \textit{xy}-plane images are of much higher resolution, a low-resolution counterpart can be generated either by manual downsampling~\cite{emdiffuse} or by learning a degradation model~\cite{cycleganir}, forming paired HR–LR supervision for training; during inference, the resulting super-resolution model is then applied along the \textit{z}-axis to enhance \textit{xz/yz} slices.
Both paradigms implicitly assume spatial isotropy. In contrast, our method directly performs inference in continuous 3D space, without relying on such isotropic assumptions.
}
\vspace{-18pt}
\label{fig:intro}
\end{figure}

Deep learning has emerged as a powerful tool to address this anisotropy by computationally enhancing axial resolution~\cite{heinrich2017deep,emdiffuse,cycleganir,hagita2018super,he2023isovem}. These methods aim to restore high-resolution details from inexpensive anisotropic acquisitions, thereby boosting imaging throughput while preserving structural fidelity.
Conventional learning-based approaches typically rely on supervised training using paired high- and low-resolution volumes~\cite{heinrich2017deep}, or on partially isotropic samples~\cite{emdiffuse}, leading to strong dependence on well-curated data and poor adaptability across different domains and imaging conditions.
In practice, isotropic reference data are rarely available, as the need for isotropic reconstruction mostly arises from the absence of such acquisitions.
Self-supervised alternatives have been explored to alleviate the data reliance on isotropic data, including GAN-based~\cite{cycleganir,hagita2018super} and diffusion-based~\cite{pan2023diffuseir,emdiffuse} methods.
Their key insight is to leverage the high-resolution \textit{xy}-planes as an internal reference.
As shown in Figure~\ref{fig:intro}, to reconstruct a volume from anisotropic slices, they either perform \textit{video frame-interpolation} between consecutive \textit{xy}-slices to generate intermediate frames~\cite{cycleganir,he2023isovem}, or apply \textit{image super-resolution} to orthogonal views (e.g., \textit{xz} or \textit{yz}) to improve resolution along the \textit{z}-axis~\cite{emdiffuse}. 
Both strategies rely on the critical assumption that local tissue structures are approximately isotropic across the \textit{x}, \textit{y}, and \textit{z} dimensions, such that geometric patterns in the high-resolution \emph{xy}-planes can be extrapolated along the axial direction.

In practice, however, the assumption of isotropy is often violated in real biological specimens, where morphological anisotropy is prevalent---e.g., elongated neuronal fibers or dendritic spines exhibit strong directional continuity~\cite{zhang2012noddi,beaulieu2002basis}.
Consequently, these methods suffer from limitations in accurately segmenting or analyzing such structures. This motivates the need for frameworks that move beyond 2D image priors to directly reason about 3D space and structural continuity, capable of preserving the true morphology of biological specimens.

In this paper, we introduce \model{}, a novel framework that reformulates anisotropic volume reconstruction as a \textit{dynamic 3D scene rendering} problem via dynamic Gaussian splatting. 
Instead of processing individual slices separately, we represent the entire axial slice sequence as the temporal evolution of a dynamic 2D Gaussian point cloud. 
By enriching each Gaussian with dynamic attributes that allow deformation, spatial shifting, and opacity variation across adjacent slices, our model implicitly recovers the underlying continuous 3D structure and synthesizes intermediate slices at arbitrary depths without relying on large models or external datasets.
To further improve fidelity under sparse supervision, we employ a \textit{teacher-student pseudo-labeling scheme} in which an EMA-based teacher provides stable pseudo targets for unseen slices, progressively guiding the student toward smoother and more consistent interpolation along the axial dimension.
This strategy effectively regularizes the network against overfitting to the limited observed slices and promotes smooth, geometry-aware interpolation across the volume.

In this way, our method fundamentally circumvents the domain mismatch issue inherent in prior isotropic-assumed models. Moreover, \model{} offers two additional advantages.
First, unlike methods such as EMDiffuse~\cite{emdiffuse}, which require additional isotropic sub-volumes of the same tissue for training, our approach operates in a fully self-contained optimization loop that uses only the anisotropic slices from the target volume. This makes it suitable for data-scarce or computationally constrained scenarios without the need for large-scale pretraining or auxiliary datasets. 
Second, by explicitly modeling geometric and photometric evolution across slices, our method enforces spatial continuity and physical coherence in the reconstructed volume, leading to stable training and reduced artifacts in the final reconstructed volume.

Extensive experiments on multiple vEM datasets demonstrate that \model{} achieves lower test-time reconstruction error and delivers superior visual quality compared to existing methods, producing more realistic details with fewer artifacts. 
By providing a stable and highly generalizable solution, our framework not only advances the state of the art in vEM reconstruction but also establishes a scalable paradigm for slice-to-3D reconstruction across diverse imaging modalities.

\section{Related Work}
\label{sec:related}

\subsection{Isotropic Reconstruction of Volume EM}

Volume Electron Microscopy (vEM) refers to a suite of techniques for generating 3D volumes of biological tissues, which are foundational for analyzing complex 3D cellular structures~\cite{li2025situ,glausier2025volume,turegano2024tracing,zhao2024application}. 
A standard vEM workflow consists of sequentially sectioning and imaging a specimen, then computationally aligning and reconstructing them into a coherent 3D volume~\cite{peddie2022volume,collinson2023volume}. 
However, achieving isotropic reconstruction for large volumes remains elusive, as high-throughput methods like ssTEM~\cite{briggman2006towards},ssSEM~\cite{hayworth2014imaging} and SBF-SEM~\cite{denk2004serial} prioritize speed over Z-axis fidelity, while isotropic techniques like enhanced FIB-SEM~\cite{bushby2011imaging} are limited to small volumes and require specialized instrumentation~\cite{peddie2022volume}.
Thus, achieving isotropic reconstruction for biological specimens remains a persistent challenge.

Early approaches addressed anisotropic reconstruction through slice interpolation or optical-flow-based alignment~\cite{carata2011improving,gonzalez2022optical,ferede2025z,joshi2025interpolai}. While straightforward to implement, these methods often introduce severe artifacts under high anisotropy factors.
Deep learning approaches initially relied on fully supervised training. 
For instance, 3DSRUNet~\cite{heinrich2017deep} learns interpolation from paired high- and low-resolution volumes and achieves strong performance on synthetic data. 
However, acquiring such paired isotropic/anisotropic volumes from the same tissue requires specialized imaging protocols, such as dual-mode acquisition or repeated scans, which are rarely feasible for most vEM systems~\cite{peddie2022volume}. 
Moreover, synthetic degradations (e.g., row removal or Gaussian blur) poorly approximate real imaging processes, leading to domain gaps and limited generalization.
These limitations motivate self-supervised approaches that avoid the need for isotropic training data. 

Early self-supervised methods~\cite{weigert2018content,krull2019noise2void} construct pseudo HR–LR pairs internally, for example by masking pixels or exploiting axis re-interpretation. 
However, they typically assume a known point spread function (PSF), whereas the true degradation in practice is often unknown and deviates from handcrafted models.
Alternative strategies such as CycleGAN~\cite{cycleganir} attempt to learn the degradation process directly from target tissue data. 
However, adversarial training can introduce instability and artifacts, including texture aliasing and geometric distortion.
More recently, diffusion-based models~\cite{emdiffuse,pan2023diffuseir} learn generative priors from abundant \textit{xy}-plane images and leverage them to enhance axial resolution. 
Although these approaches improve fidelity and stability compared with GAN-based methods, they operate in a slice-wise 2D manner and implicitly assume spatial isotropy. 
As a result, degradations learned in the lateral (\textit{xy}) plane may not transfer well to the axial (\textit{z}) direction.
In contrast, our method directly models continuous 3D geometry across slices, avoiding isotropy assumptions and enabling more faithful reconstruction of anisotropic structures.

\subsection{3D Gaussian Splatting and Dynamic Variants}

Recent advances in neural rendering have been significantly driven by the emergence of 3D Gaussian Splatting (3DGS) and its extensions to dynamic scenes.
3DGS~\cite{kerbl20233d} represents a scene as a set of anisotropic 3D Gaussians parameterized by position, covariance, opacity, and spherical harmonic coefficients for view-dependent appearance. This explicit representation enables efficient optimization and high-quality real-time rendering, substantially improving both speed and fidelity compared with earlier explicit~\cite{muller2022instant} and implicit~\cite{mildenhall2021nerf} scene representations.

To handle dynamic scenes, subsequent works introduce temporal modeling through deformation fields that map canonical Gaussians to time-dependent observations. 
Deformable 3D Gaussians~\cite{yang2024deformable}, for example, employ a lightweight multi-layer perceptron (MLP) to predict per-Gaussian deformations conditioned on timestamps, enabling reconstruction of non-rigid motions from monocular videos. 
Similarly, 4D Gaussian Splatting (4DGS)~\cite{wu20244d} models motion using a 4D neural voxel representation decomposed with a HexPlane-style encoding, followed by a compact MLP decoder for efficient deformation prediction. 
These approaches extend the efficiency of Gaussian primitives to dynamic scene reconstruction while explicitly modeling temporal variation.

Beyond volumetric rendering, recent studies also explore Gaussian splatting in the 2D image domain for tasks such as image restoration, compression, and super-resolution~\cite{zhang20242d,zhang2024gaussianimage,hu2025gaussiansr,zhu2025large}. 
These works highlight the flexibility of Gaussian primitives and suggest their broader applicability to imaging problems beyond 3D scene reconstruction.

\begin{figure}[t]
\centering
\includegraphics[width=\linewidth]{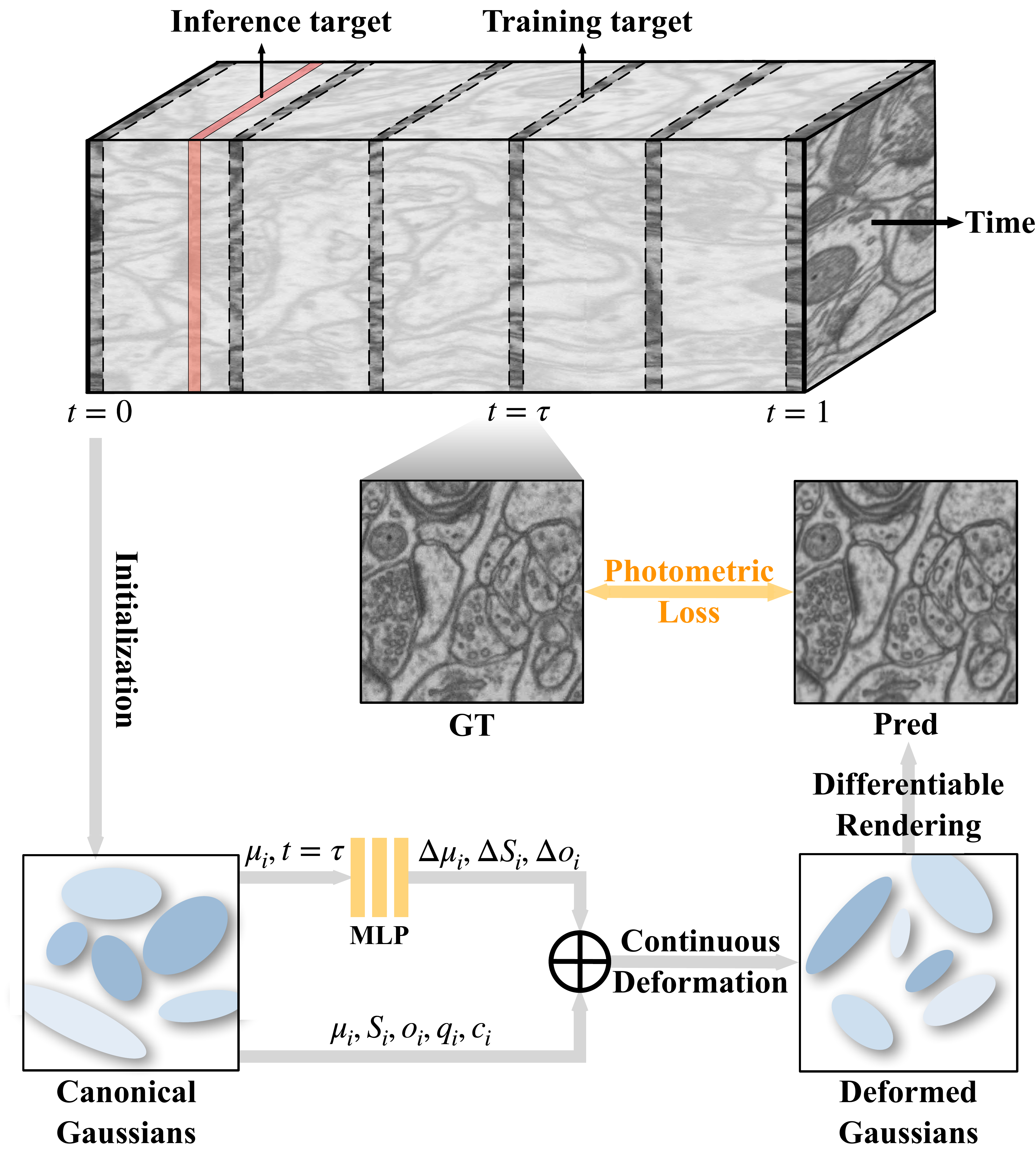}
\vspace{-12pt}
\caption{\textit{Overview of the pipeline of \model{}}. 
Given a volume EM dataset (top), the gray-shaded regions denote the missing intermediate slices to be interpolated, while the visible regions represent the available anisotropic training data. 
We interpret the low-resolution $z$-axis as a temporal dimension and employ a \textbf{deformable 2D Gaussian splatting} representation to model the geometric evolution of 2D structures along depth. 
During inference, the model synthesizes the target slice by directly conditioning on the relative temporal coordinate corresponding to the desired interpolation position.
}
\vspace{-12pt}
\label{fig:methods}
\end{figure}

\section{Method}
\label{sec:method}

\subsection{Preliminaries}

Our method builds upon Deformable 3D Gaussians~\cite{yang2024deformable},an extension of 3D Gaussian Splatting (3DGS)~\cite{kerbl20233d} for dynamic representations.
3DGS models a scene as a collection of explicit 3D Gaussian primitives, each parameterized by an opacity $o\in[0,1]$, a center $\mu\in\mathbb{R}^{3\times1}$, and a covariance matrix $\Sigma\in\mathbb{R}^{3\times3}$:
\begin{equation}
G(\mathbf X)=\exp\left[-\tfrac{1}{2}(\mathbf X-\mu)^{\top}\Sigma^{-1}(\mathbf X-\mu)\right].
\end{equation}
The covariance $\Sigma$ is typically decomposed as $\Sigma=R S S^{\top}R^{\top}$, where $S\in\mathbb{R}^{3\times1}$ denotes the scaling factors and $R\in\mathbb{R}^{3\times3}$ is a quaternion-parameterized rotation matrix, enabling efficient gradient-based optimization.
For image rendering from a given viewpoint, Gaussians within the camera frustum are projected to the image plane via a splatting process. The projected 2D covariance $\Sigma'$ is given by $\Sigma' = J W \Sigma W^{\top} J^{\top}$, where $W$ and $J$ denote the view transformation and the Jacobian of the local projective mapping, respectively. Each pixel color $\mathbf p$ is then obtained by alpha-blending the overlapping Gaussians ${G_i}_{i=1}^N$ in front-to-back order.

Building on this foundation, Deformable 3D Gaussians~\cite{yang2024deformable} extend 3DGS to dynamic scenes by learning a canonical Gaussian set $\mathcal{G}c = {(\mu_i, S_i, q_i, o_i, \mathcal{C}i)}_{i=1}^M$ along with a lightweight MLP-based deformation network $\Phi\theta$.
Conditioned on a temporal coordinate $t\in[0,1]$, $\Phi_\theta$ predicts per-Gaussian offsets:
\begin{equation}
\Delta \mu_i,, \Delta S_i,, \Delta q_i,, \Delta o_i = \Phi_\theta(\mu_i, t),
\end{equation}
The deformed Gaussian at time $t$ is then obtained by applying these deltas to each attribute correspondingly. 
This formulation enables smooth temporal interpolation while preserving fine-grained spatial details.

\begin{figure*}[t]
  \centering
  \includegraphics[width=\linewidth]{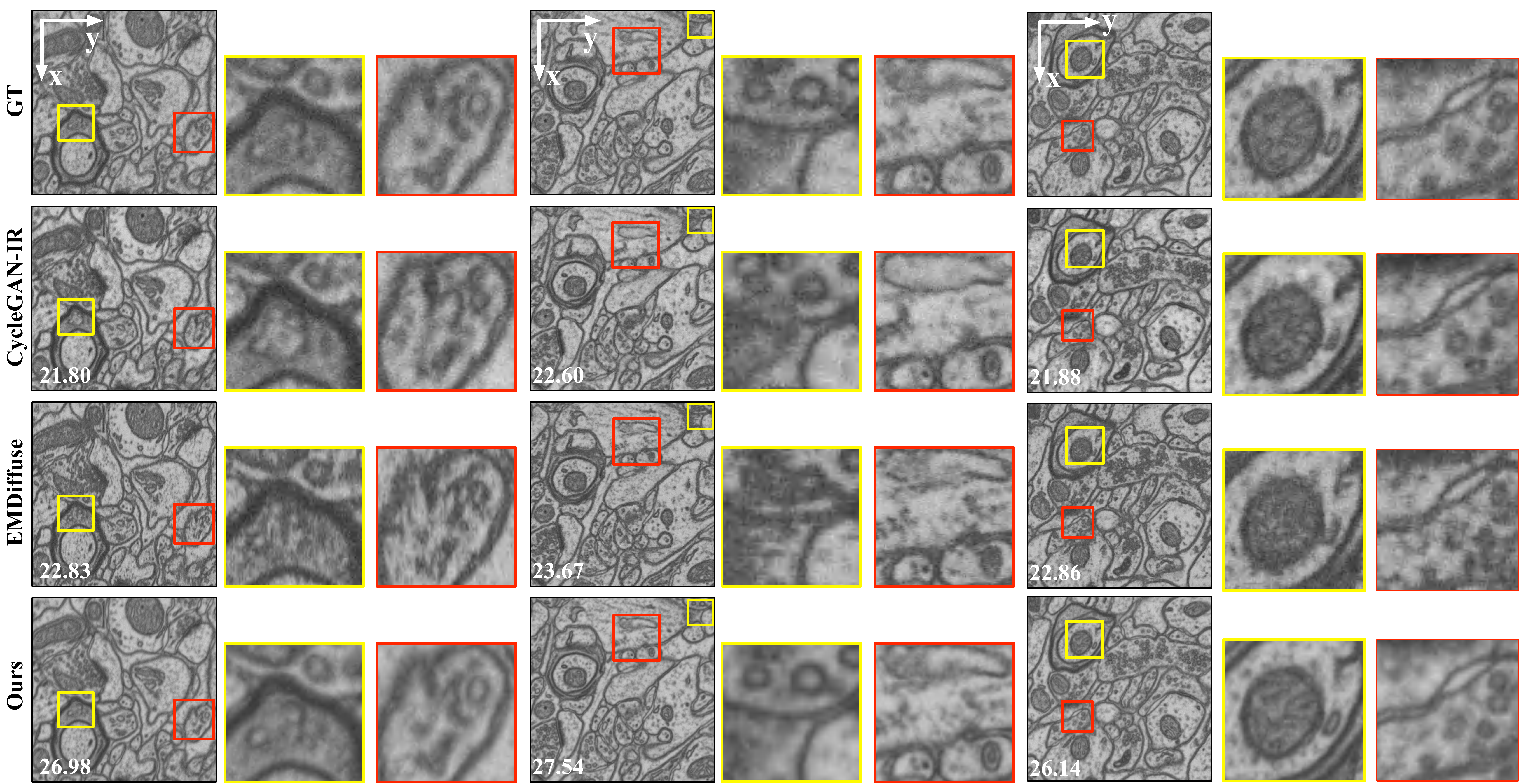}
   \caption{\textit{Isotropic xy-slice reconstruction results on EPFL dataset~\cite{lucchi2013learning}.}}
   \vspace{-12pt}
   \label{fig:iso}
\end{figure*}

\subsection{Slice-to-3D Dynamic Gaussians}

Given a set of anisotropic slices ${I_t}$ sparsely sampled along the z-axis, our goal is to reconstruct a continuous isotropic volume $\mathcal{V}$ by learning a temporally continuous deformation field that bridges adjacent slices.
Unlike dynamic 3D reconstruction tasks, which deform Gaussians over time for 3D motion modeling, our deformation field is designed to parameterize spatial continuity along the axial (z) dimension.
Specifically, we treat the slice index $t \in [0,1]$ as a normalized spatial coordinate along the z-axis, and train a deformation MLP to learn the local geometric variations between neighboring slices.

As shown in Figure \ref{fig:methods}, starting from a canonical Gaussian set $\mathcal{G}c={(\mu_i,S_i,R_i,o_i,\mathcal{C}i)}_{i=1}^M$ initially optimized from the observed anisotropic frames, the deformation network $\Phi_\theta$ predicts local offsets for each Gaussian as:
\begin{equation}
\Delta\mu_i,\ \Delta S_i,\ \Delta o_i = \Phi_\theta(\mu_i, t),
\end{equation}
where $\Delta\mu_i = (\Delta x_i, \Delta y_i, 0)$ encodes lateral displacements, $\Delta S_i = (\Delta s_x, \Delta s_y, 0)$ controls in-plane scaling, and $\Delta o_i$ modulates the opacity change along $z$.

To maintain training stability and prevent degenerate deformation along the depth axis, we fix both the absolute z-coordinate and the z-scale of all Gaussians to shared constants, while allowing rotation to be learned but not time-dependent. 
Formally:
\begin{align}
z_i &= z_0, \\
s_{z,i} &= s_{z,0}, \\
\Delta z_i &= \Delta s_z = \Delta R_i = 0,
\end{align}
where $z_0$ and $s_{z,0}$ are global constants shared across all Gaussians. 
This ensures consistent axial alignment while enabling learnable in-plane deformation and per-slice appearance variation.

During inference, we query $\Phi_\theta$ with intermediate values of $t$ to generate deformed Gaussian sets $\mathcal{G}_t$, which are then rendered via Gaussian splatting to produce interpolated slices $\hat{I}_t$.
This process enables continuous z-axis upsampling without requiring any additional training data beyond the given anisotropic volume.

\begin{algorithm}[t]
    \caption{EMA-based Teacher–Student Bootstrapping.}
    \begin{algorithmic}[1]
    \small
    \State \textbf{Input: } Training set $ \{\mathcal D_i\}_{1:N}$, where $ D_t= \{$image $I_i$, timestamp $t_i\in[0,1]\}$
    \State \textbf{Hyperparameters:} Learning rates $\eta$, EMA decay $\alpha$, Loss weight $\lambda_\text{pseudo}$
    \State \textbf{Parameters:} 
    Gaussian attributes $\{\mathcal{G}_j\}_{1:M}$ where $ G_j= (\mu_i,S_i,R_i,o_i,\mathcal{C}i)$,
    Pretrained deformation network $\Phi_{\theta_\text{student}}$
    \State Initialize teacher network $\Phi_{\theta_\text{teacher}} \leftarrow \Phi_{\theta_\text{student}}$
    \While{not converged}
        \State Sample an unseen timestamp $\tilde{t} \notin \{\mathcal D_i\}_{1:N}$
        \State $\mathcal{L}_\text{pseudo} = \lambda_\text{pseudo}\|\Phi_{\theta_\text{student}}(\{\mathcal{G}_j\}, \tilde{t}) - \Phi_{\theta_\text{teacher}}(\{\mathcal{G}_j\}, \tilde{t})\|$
        \State $\theta_\text{student} \leftarrow \theta_\text{student} - \eta\nabla_{\theta_\text{student}}\mathcal{L}_\text{pseudo}$
        \State $\theta_\text{teacher} \leftarrow \alpha \cdot \theta_\text{teacher} + (1-\alpha)\cdot\theta_\text{student}$
    \EndWhile
    \end{algorithmic}
    \label{teacher}
\end{algorithm}

\subsection{Bootstrapping with Pseudo Gaussian Targets}

To further enhance the fidelity of interpolated slices in data-sparse regimes, we introduce a \textit{Teacher-Student pseudo-labeling framework} that leverages high-confidence predictions on unobserved intermediate slices as supervisory signals. 
It can be viewed as a bootstrapping method as we use slice predictions to generate new slice predictions.
This approach is particularly effective in our setting, where only a small subset of axial slices (e.g., $10\%-20\%$) are provided during training, leaving the majority of the \textit{z}-axis unobserved.

As shown in Algorithm \ref{teacher}, we maintain a stable teacher model as an exponential moving average (EMA) version of the student model (the actively trained model). The teacher parameters $\theta_{\text{teacher}}$ are updated at each optimization step (Line 9), where $\alpha = 0.995$ is the EMA decay rate. 
Concretely, the student is trained with a matching pseudo-label loss on selected unseen slices, which is detailed in Section \ref{pipeline}.
The teacher model is initialized when the training iteration reaches a predefined threshold, ensuring initial convergence before introducing pseudo-supervision. After a short warm-up, we activate the EMA teacher to generate pseudo targets on a subset of unseen slices and gradually include more unseen indices as training progresses (from mid-slices to other interleaving positions).
Note that the algorithm describes the pseudo-supervision iterations, where the student is optimized using the teacher’s predictions on non-training frames. In practice, these pseudo-supervised iterations are interleaved with standard iterations using ground-truth training data, ensuring balanced learning and stable convergence.

This Teacher-Student framework enables robust learning of the deformation field by providing stable pseudo-supervision for unobserved intermediate slices, effectively regularizing the model against overfitting to sparse training data while facilitating smooth interpolation along the axial dimension.

\begin{figure*}[t]
  \centering
  \includegraphics[width=\linewidth]{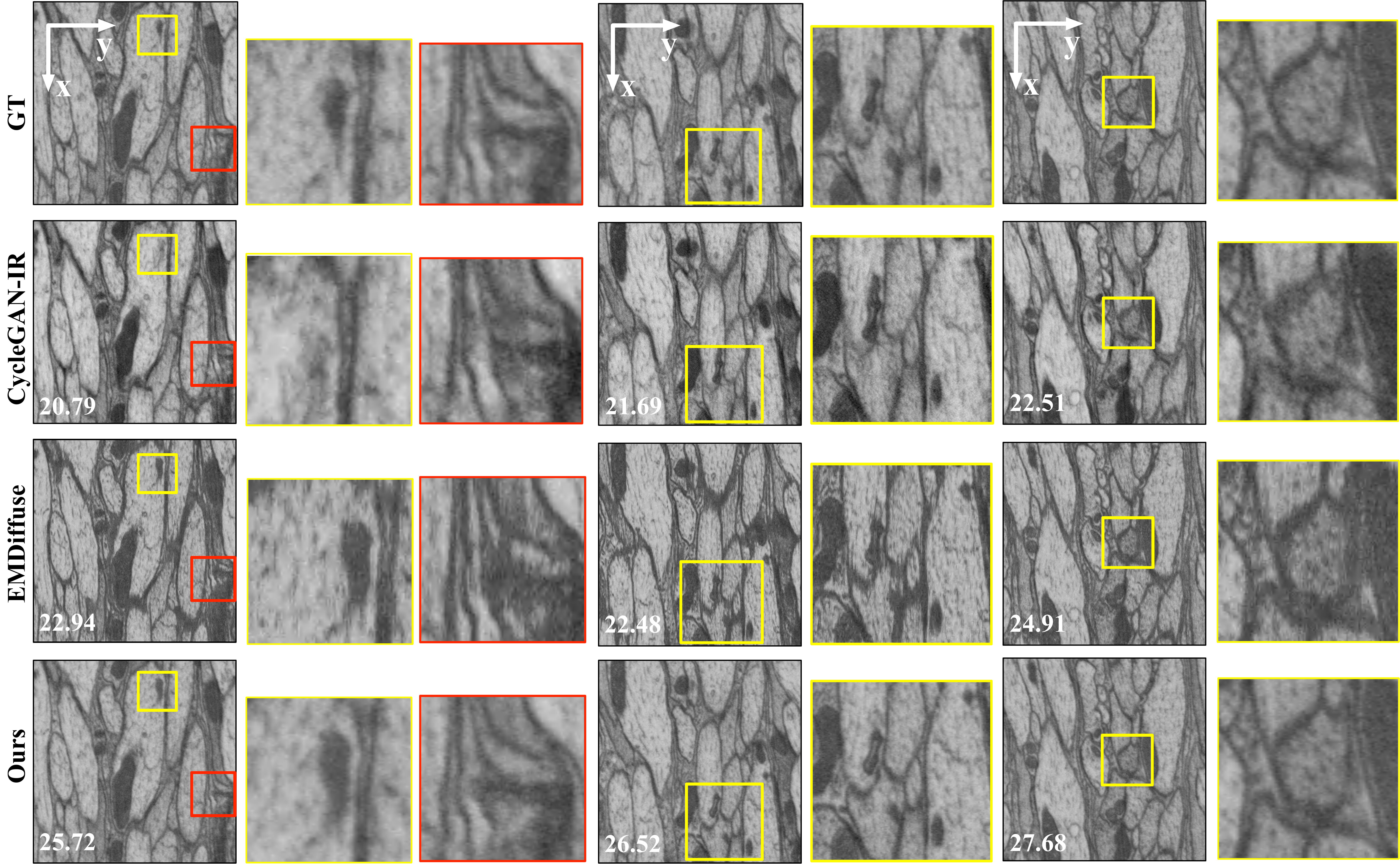}
   \caption{\textit{Isotropic xy-slice reconstruction results on FIB-25 dataset~\cite{takemura2015synaptic}.}}
   \vspace{-5pt}
   \label{fig:iso_fib}
\end{figure*}

\subsection{Training Pipeline}
\label{pipeline}

We adopt a three-stage training pipeline.
First, we optimize the canonical Gaussian set $\mathcal{G}_c$ and the rendering parameters on the observed slices $\mathcal{T}$ while freezing the deformation MLP, establishing a stable radiance baseline.
Next, we enable the deformation MLP $\Phi_\theta$ and jointly train it with $\mathcal{G}_c$ on $\mathcal{T}$ so that in-plane position offsets $(\Delta x,\Delta y)$, in-plane scales offsets $(\Delta s_x,\Delta s_y)$, and opacity offsets capture the observed axial transitions. This second stage is deliberately brief: extending it tends to overfit and trap the model in suboptimal local minima, yet omitting it leads to an initially weak teacher that hinders convergence.
Finally, we activate the EMA teacher to generate pseudo targets on a progressively enlarged subset of unseen timestamps. Training alternates between supervised iterations using ground-truth samples and pseudo-supervised iterations with teacher-generated labels, gradually increasing the proportion of pseudo-supervised steps as the model stabilizes.

\vspace{-12pt}
\paragraph{Objective functions.}
Our model can effectively function with only RGB photometric supervision. 
Following 3DGS~\cite{kerbl20233d}, we compute the deviation between the rendered image and the ground-truth RGB frame using an $\ell_1$ loss combined with a D-SSIM regularization term.
For pseudo-supervised iterations, we apply the same objective but scale the loss weight $\lambda_\text{pseudo}$ by an exponentially increasing factor, gradually enhancing the impact of pseudo supervision as training progresses.

\vspace{-12pt}
\paragraph{Implementation details.}
We implement \model{} in PyTorch~\cite{paszke2019pytorch} and optimize with Adam~\cite{kingma2014adam} following the default hyperparameters of 3DGS~\cite{kerbl20233d} for Gaussian attributes and learning rates. 
Training proceeds in three phases: 
a warm-up phase of $2$k iterations optimizing the canonical Gaussian set $\mathcal{G}_c$ with the deformation MLP frozen, a short joint-training phase of $1$k iterations enabling the $\mathcal{G}_c$ and the deformation MLP, and a final phase of $15k$ iterations with the EMA teacher providing pseudo supervision on unseen timestamps until convergence.
During the pseudo-supervised phase, the corresponding loss weight is linearly ramped from $0.1$ to $1.0$ between $3$k and $10$k iterations.
The EMA teacher employs a decay factor of $\alpha = 0.995$.
All experiments are conducted on an NVIDIA RTX~3090 GPU.

\section{Results}
\label{sec:results}

\subsection{Experimental Settings}

\paragraph{Datasets.}
We conduct experiments on diverse large-scale electron microscopy (EM) datasets under both isotropic and anisotropic imaging conditions, as shown in Table~\ref{tab:dataset}. For isotropic data, we select the \textit{EPFL mouse brain dataset}~\cite{lucchi2013learning} acquired from the CA1 hippocampus region of the mouse brain with a uniform voxel resolution of $5\times5\times5$ nm, and the \textit{FIB-25 drosophila brain dataset}~\cite{takemura2015synaptic} with a resolution of $8\times8\times8$ nm. Both datasets are collected by FIB-SEM. To simulate anisotropic acquisition, we subsample the $z$-axis of these isotropic volumes by a factor of $\times6$, using the withheld intermediate slices as ground truth for evaluation. Following CycleGAN-IR~\cite{cycleganir}, we use a $500\times500\times500$ sub-volume from each dataset. 
To further assess performance under real-world anisotropic imaging, we use the \textit{FANC dataset}~\cite{phelps2021reconstruction} acquired from the ventral nerve cord of an adult female Drosophila with an anisotropy ratio of $10$ ($4\times4\times40$ nm), where a $500\times500\times50$ sub-volume is used for experiments. These datasets collectively span multiple organisms, tissue types, and imaging resolutions, providing a robust benchmark for both simulated and real isotropic reconstruction. We use PSNR, SSIM~\cite{wang2004image}, and FSIM~\cite{zhang2011fsim} for quantitative evaluation.

\vspace{-10pt}
\paragraph{Baselines.}
We compare our method with representative isotropic reconstruction approaches spanning GAN-based, diffusion-based, and transformer-based paradigms, including vEMDiffuse-a from EMDiffuse~\cite{emdiffuse}, CycleGAN-IR~\cite{cycleganir}, and IsoVEM~\cite{he2023isovem}.
vEMDiffuse-a is a diffusion-based model that learns anisotropic-to-isotropic mappings using additional training data from the same tissue type.
% vEMDiffuse-a is a diffusion-based model that requires additional training data from the same tissue type beyond the target dataset to learn anisotropic-to-isotropic mapping. To train the model, we use another volume from the same tissue as the target volume, with the same spatial size, as the training data.
%
CycleGAN-IR is an unsupervised GAN-based framework built on CycleGAN~\cite{zhu2017unpaired} that learns cross-domain mappings between high-resolution lateral and low-resolution axial slices.
% CycleGAN-IR~\cite{cycleganir} is an unsupervised framework using a CycleGAN~\cite{zhu2017unpaired} architecture that learns cross-domain mappings between unpaired high-resolution (HR) lateral and low-resolution (LR) axial slices through cycle-consistent adversarial training.
%
IsoVEM~\cite{he2023isovem} is a transformer-based approach designed for anisotropic EM volume reconstruction.
%
% Table~\ref{tab:cost} further compares the training cost required by each method to generate one reconstructed sample.
% 所有baseline都需要较大规模数据集来进行长时间的预训练，whereas
% To effectively generate the in-between slices, we only need a small sub-batch of the overall data, achieving the minimum requirement on data amount and time consumption.
% All baseline methods require large-scale datasets and extensive training to achieve satisfactory performance. In contrast, our method operates with a small sub-batch sampled from the target volume and can directly generate intermediate slices, significantly reducing both data requirements and training time.
All baselines rely on large-scale datasets and extensive training, whereas our method generates intermediate slices using only a small sub-batch from the target volume, greatly reducing data and training cost.

\subsection{Isotropic Reconstruction for Simulated\\Anisotropic vEM Data}
\textbf{Results on xy-slices.} We evaluate all the methods on the xy-slice images, as shown in Figure~\ref{fig:iso}, Figure \ref{fig:iso_fib}, and Table~\ref{tab:iso}. In Figure \ref{fig:iso} and Figure \ref{fig:iso_fib}, compared with CycleGAN-IR and EMDiffuse, our method can generate more high-quality images that are closer to the GT images. The generated images from CycleGAN-IR contain some structures that do not belong to the GT images, which not only leads to a loss in image metrics but also more easily affects the results of subsequent 3D reconstruction. The images generated by EMDiffuse exhibit relatively obvious artifacts, with details and structures significantly differing from the ground truth images. This is because the training and testing domains of this method are different: the training set used downsampled xz/yz-slice images, causing the structural details on the test set to resemble those of the downsampled images. This severely degrades the quality of the generated images and poses difficulties for subsequent 3D reconstruction. Our method avoids the use of xz/yz slices and instead performs continuous modeling directly on xy-slices, thereby avoiding the problem of differing training and testing domains. Compared with the other two baseline methods, the images generated by our method contain neither extraneous structures nor blurry details, thus achieving higher image metrics and facilitating the downstream 3D reconstruction process. 

\begin{table}[t]
\caption{\textit{Dataset description.} We use both isotropic and anisotropic datasets for our experiments. For isotropic datasets, we simulate anisotropies via manual sub-sampling. }
\vspace{-15pt}
% \small
\begin{center}
\resizebox{\linewidth}{!}{
\begin{tabular}{lccccccccc}
    \toprule
    Datasets & & Isotropy & Resolution (nm) & Anisotropy ratio & Issue \\
    \midrule
    EPFL~\cite{lucchi2013learning} &   &$\checkmark$ & $5 \times 5 \times 5$ & $6\times$ & brain	       \\
    FIB-25~\cite{takemura2015synaptic}  	  &   &$\checkmark$& $8 \times 8 \times 8$ & $6 \times$  & brain   \\
    FANC~\cite{phelps2021reconstruction}  	  &   &$\times$& $4 \times 4 \times 40$ & $10 \times$& nerve     \\
    \bottomrule
\end{tabular}
}
\end{center}
\vspace{-10pt}
\label{tab:dataset}
\end{table}

\begin{table}[t]
\caption{\textit{Results of isotropic xy-slice reconstruction on synthetic anisotropic datasets.}
$^{\dagger}$EMDiffuse uses additional data for training.}
\vspace{-15pt}
% \small
\begin{center}
\resizebox{\columnwidth}{!}{
\begin{tabular}{lcccccc}
    \toprule
    \multirow{2.5}{*}{Method} & \multicolumn{3}{c}{EPFL}  & \multicolumn{3}{c}{FIB-25}\\
    \cmidrule(lr){2-4}  \cmidrule(lr){5-7} 
    & PSNR & SSIM & FSIM & PSNR & SSIM  & FSIM  \\
    \midrule
    % Cubic Interp     &   24.54&0.544&0.913&0.220     &   27.63&0.681&0.912&0.221       \\
    % EMDiffuse\cite{emdiffuse}$^{\dagger}$   &   23.34&0.519&0.899&0.298     &   24.10&0.514&0.878&0.371       \\
    % CycleGAN-IR\cite{cycleganir}            &   22.05&0.491&0.856&0.282	    &   22.39&0.554&0.856&0.298       \\
    CycleGAN-IR\cite{cycleganir}            &   22.05&0.491&0.856	    &   22.39&0.554&0.856       \\
    EMDiffuse\cite{emdiffuse}$^{\dagger}$   &   23.34&0.519&0.899       &   24.10&0.514&0.878       \\
    IsoVEM~\cite{he2023isovem}              &   23.91&0.597&0.8558     &   21.51&0.546&0.8456       \\
    Ours  	  &   \textbf{26.59}&\textbf{0.6977}&\textbf{0.943}   &   \textbf{27.37}&\textbf{0.7275}&\textbf{0.920}     \\
    \bottomrule
\end{tabular}
}
\end{center}
\vspace{-10pt}
\label{tab:iso}
\end{table}

\begin{figure}[t]
  \centering
  \includegraphics[width=\linewidth]{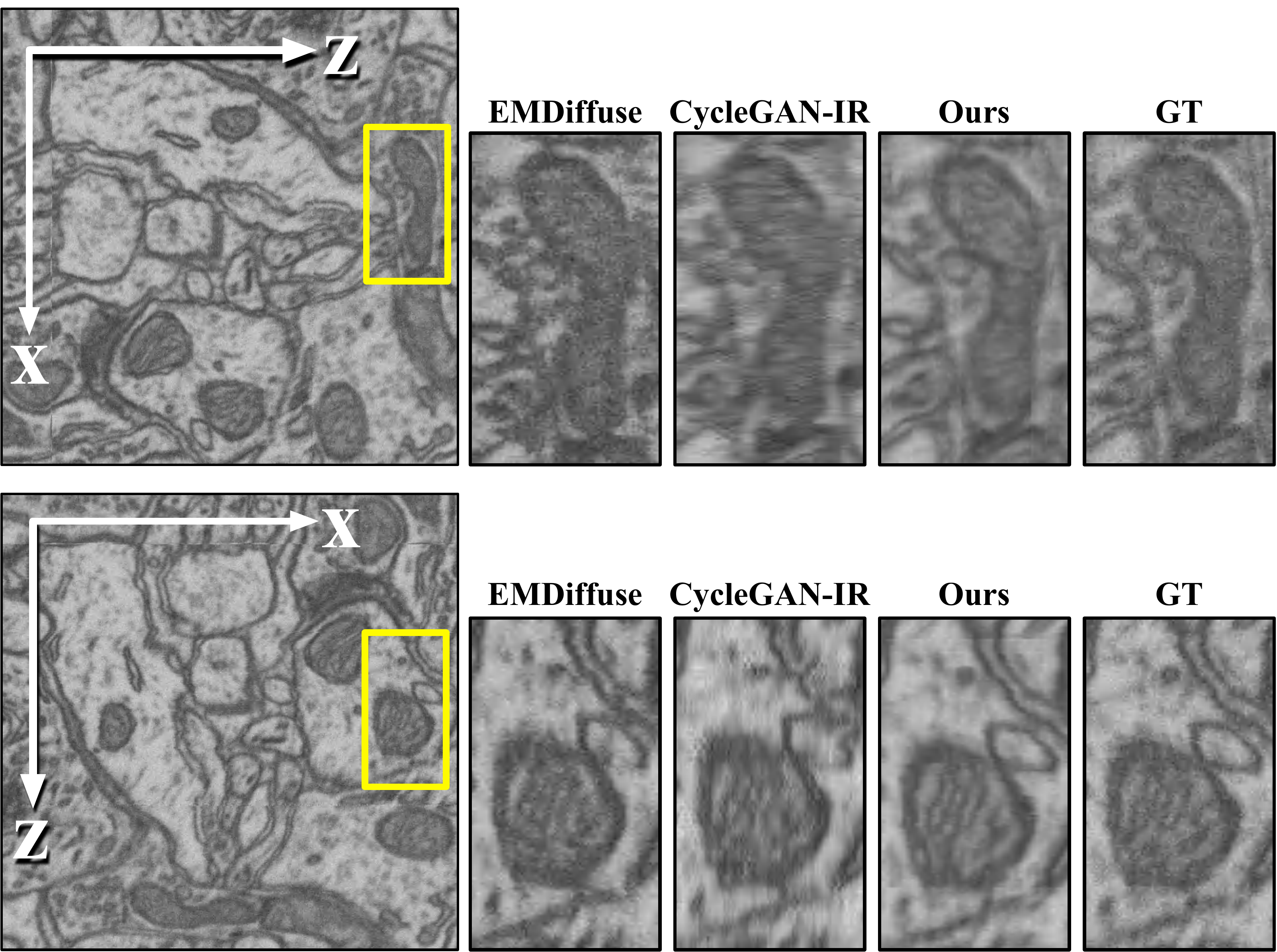}
  \vspace{-15pt}
   \caption{\textit{Isotropic xz/yz-slice Reconstruction results on synthetic anisotropic datasets.}}
   \vspace{-10pt}
   \label{fig:iso_xz}
\end{figure}

\vspace{-10pt}
\paragraph{Results on xz/yz-slices.} 
We further evaluate all methods on reconstructed \textit{xz}/\textit{yz} slices, as shown in Figure~\ref{fig:iso_xz}. Although EMDiffuse~\cite{emdiffuse} and CycleGAN-IR~\cite{cycleganir} are trained using \textit{xz}/\textit{yz} slices, the generated results still exhibit noticeable artifacts and structural inconsistencies. This suggests that the downsampled \textit{xz}/\textit{yz} slices differ substantially from the original slices, making it difficult to accurately model the underlying volumetric structure. In contrast, although our method is trained only on \textit{xy} slices, the proposed 3D continuous deformation field enables the model to capture coherent volumetric deformation across the entire volume. As a result, our method produces smoother and more realistic reconstructions on the \textit{xz}/\textit{yz} views.
% We also test all the methods on xz/yz-slices, as shown in Figure \ref{fig:iso_xz}. It can be seen that, though trained on the xz/yz-slices, EMDiffuse and CycleGAN-IR still generate xz/yz-slices with significant artifacts. This indicates that the downsampled xz/yz-slices have significant differences from the original slice, thus methods based on it are not able to generate satisfying results. Instead, though our method is trained directly on xy-slices, thanks to the 3D continuous deformation field, our model is capable of modeling the continuous deformation of the volume, and therefore can generate smoother and realistic results.

% \begin{figure*}[h]
%   \centering
%   \includegraphics[width=\linewidth]{fig/sequence.pdf}
%    \vspace{-5pt}
%    \caption{\textit{Isotropic Reconstruction results on EPFL datasets with anisotropic factor of $6\times$.}}
%    \label{fig:sequence}
% \end{figure*}

\begin{figure}[t]
  \centering
  \includegraphics[width=\linewidth]{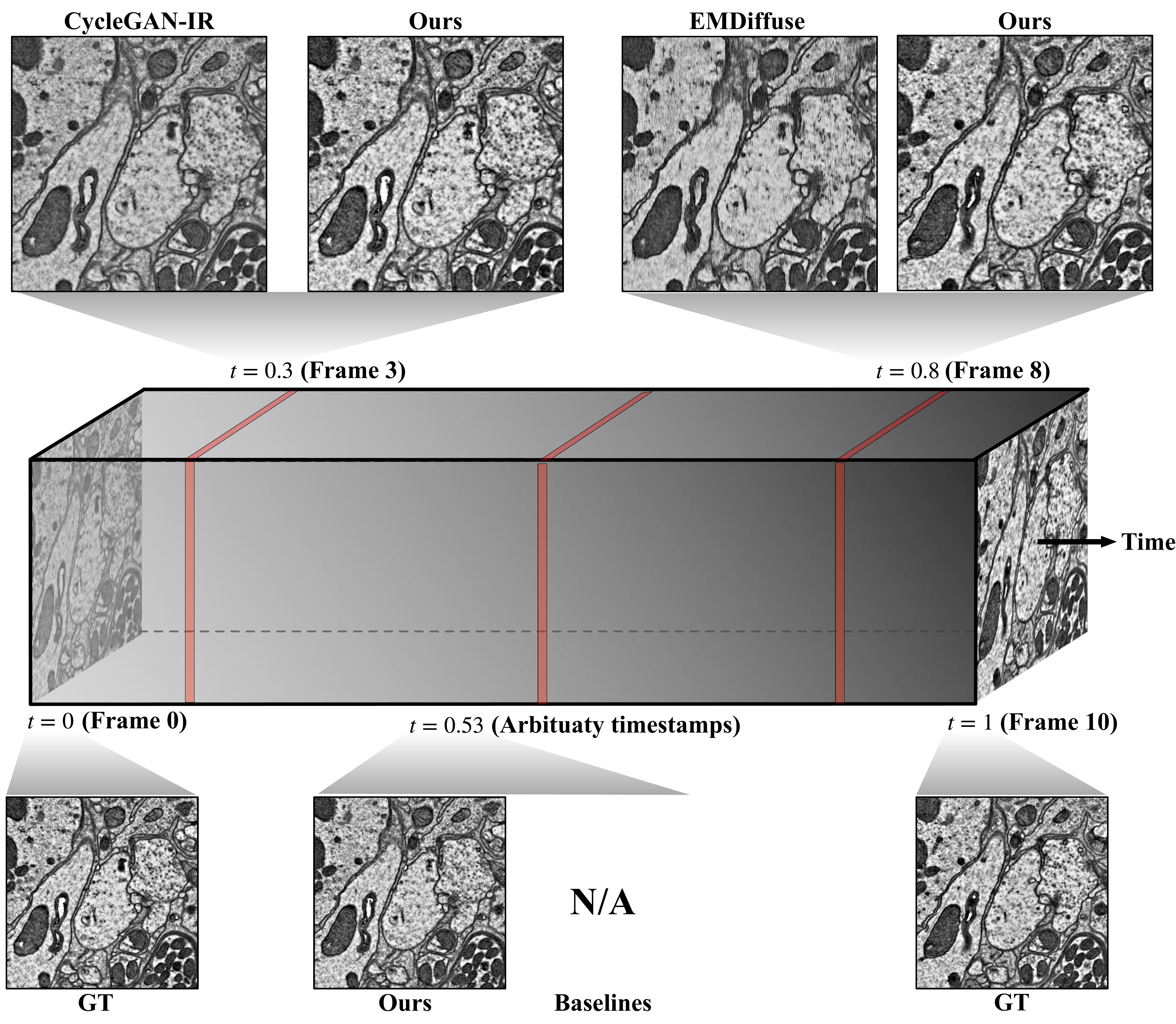}
  % \vspace{-5pt}
   \caption{\textit{Isotropic xy-slice Reconstruction results on real-captured anisotropic \textit{FANC datasets} with anisotropic factor $10\times$.} Notably, beyond the predefined slice intervals, our model can freely interpolate frames at any arbitrary time step (e.g. $t=0.53$).}
   \label{fig:aniso}
   \vspace{-8pt}
\end{figure}

\subsection{Isotropic Reconstruction for Real Anisotropic\\vEM Data}
To demonstrate the generalizability of our method to real-world scenarios, we evaluate all methods on the real anisotropic Volume EM dataset FANC~\cite{phelps2021reconstruction}. The results are shown in Figure~\ref{fig:aniso}. Compared with CycleGAN-IR~\cite{cycleganir} and EMDiffuse~\cite{emdiffuse}, our method produces more accurate and structurally detailed reconstructions, indicating that it remains effective on real anisotropic data. Moreover, only our method enables continuous generation along the $z$-axis, allowing images to be synthesized at an arbitrary timestamp $t$. As illustrated in Figure~\ref{fig:aniso}, when evaluated at an arbitrary timestamp $t=0.53$, CycleGAN-IR~\cite{cycleganir} and EMDiffuse~\cite{emdiffuse} cannot produce valid results because they rely on discrete \textit{xz}/\textit{yz} slice interpolation. In contrast, our method models the entire 3D volume and can therefore generate consistent slices at any timestamp.

% \paragraph{More results.}
% 3D Reconstruction on Anisotropic VolumeEM Dataset. 
% xz/yz images.
% Supplementary materials.

\begin{table}[t]
\caption{\textit{Video segmentation results on EPFL datasets with synthetic anisotropy ratio of $\times6$.
Segmentation is performed on the reconstructed isotropic slice sequences using SAM2~\cite{ravi2024sam2}, and IoU is computed against the ground-truth intermediate slices.
}}
\vspace{-18pt}
\footnotesize
\begin{center}
% \resizebox{\linewidth}{!}{
\begin{tabular}{lccc}
    \toprule
    method & CycleGAN-IR & EMDiffuse & Ours \\\midrule
    IoU & 0.9099&0.9555&0.9687      \\
    \bottomrule
\end{tabular}
% }
\end{center}
\vspace{-12pt}
\label{tab:iou}
\end{table}

\begin{table}[t]
\caption{\textit{Ablation studies on key components of our framework.} The results are averaged over the two isotropic datasets.}
\vspace{-15pt}
\footnotesize
\begin{center}
% \resizebox{\linewidth}{!}{
\begin{tabular}{lccc}
    \toprule
    Method & PSNR & SSIM & FSIM \\
    \midrule
    w/o teacher-student module& 25.19&0.6272&0.9035 \\
    w/o Warmup&                 25.76&0.6528&0.9076 \\    
    w/o Joint&                  24.35&0.5773&0.8513 \\
    w/o $\Delta o$&             25.44&0.6295&0.8936 \\
    w/ $\Delta R$&              25.07&0.6398&0.9055 \\
    Ours&\textbf{26.98}&\textbf{0.7126}&\textbf{0.9315} \\
    \bottomrule
\end{tabular}
% }
\end{center}
\vspace{-15pt}
\label{tab:ablation}
\end{table}

\subsection{Downstream Segmentation Results}
Isotropic slice interpolation is essential for accurate 3D reconstruction in volume electron microscopy (vEM). 
To evaluate downstream utility, we follow~\cite{emdiffuse} and perform video-based segmentation on the reconstructed volumes.
Specifically, we adopt Segment Anything Model 2 (SAM2)~\cite{ravi2024sam2} to segment cellular structures across continuous \textit{xy} slices.
With ground-truth intermediate slices available, we compute their segmentation masks and measure the Intersection-over-Union (IoU) between predictions and ground truth to assess segmentation accuracy. The results are summarized in Table~\ref{tab:iou}.
We further provide 3D reconstruction visualizations and comparisons in the \textit{supplementary video}.
These experiments collectively demonstrate that high-quality anisotropy correction is crucial for downstream 3D vEM analysis, and that our reconstruction pipeline yields more coherent structural continuity and higher morphological fidelity than existing methods.

\subsection{Ablation Study}
We conduct ablation studies to systematically evaluate the contribution of key components in our framework, including the teacher–student module, training pipeline design, and Gaussian configuration.
As shown in Table~\ref{tab:ablation}, removing the teacher-student module leads to a noticeable degradation in generation quality and structural fidelity. 
Skipping the warm-up stage results in coarse initialization and compensatory Gaussian proliferation, while omitting joint training yields low-quality pseudo-labels and suboptimal convergence.
For our Gaussian configuration, we adopt dynamic opacity with time-independent rotation. Static opacity introduces noise when modeling structural appearance and disappearance, whereas dynamic rotation causes temporal jitter.
Further ablations on different anisotropy ratios ($\times4$ and $\times8$) and the number of training slices per bundle are provided in the \textit{supplementary materials}.
\section{Discussion}
\label{sec:discussion}

% Our method is a general framework for 3D reconstruction from planar scanned 2D slices with applications in vEM. Compared with other methods, it reconstruct the 3D volume in a more principled way by directly utilizing the xy slices to construct a continuous deformation field.
% Meanwhile, it also have some limitations. A main drawback is that, the number of Gaussian splating points may largely increse in certain scenes due to the noisy input image, leading to out-of-memory errors. An effective solution is to employ a denosing network before our pipeline. 

Our method provides a general and flexible framework for reconstructing 3D structures from planar-scanned 2D slices, with a particular focus on applications in vEM. Unlike prior approaches that rely on anisotropic priors or heuristic slice-to-slice matching, our formulation reconstructs the underlying 3D volume in a more principled manner by explicitly modeling a continuous deformation field directly from the high-resolution xy slices. This design not only yields smoother inter-slice transitions but also enables a more faithful recovery of morphological variations across depth. 
Despite its advantages, our framework still exhibits several limitations. A primary drawback is that the number of Gaussian splatting primitives can grow significantly in certain challenging regions---especially when the input slices contain substantial noise. The resulting increase in Gaussian density may lead to excessive memory consumption. A practical remedy is to incorporate a lightweight denoising module before our reconstruction pipeline, which can stabilize the optimization and prevent the uncontrolled proliferation of Gaussian primitives. In addition, future extensions may explore adaptive Gaussian pruning or joint learning with image-space regularizers to further improve robustness and computational efficiency.

\section{Conclusion}
\label{sec:conclusion}

% We propose EMGauss, which enhances anisotropic volume reconstruction in volume electron microscopy by casting the task as dynamic 3D scene rendering via Gaussian splatting. Our method bypasses isotropy assumptions in prior diffusion- and GAN-based methods, serving as a more princepled strategy. Through a teacher-student bootstrapping mechanism and self-supervised optimization on target slices alone, our framework achieves continuous, artifact-reduced interpolation with enhanced structural fidelity, whiout reliance on external isotropic data. Beyond vEM, EMGauss establishes a versatile, domain-agnostic paradigm for slice-to-volume synthesis, paving the way for high-throughput, high-fidelity 3D imaging across diverse modalities.

We introduced EMGauss, a simple yet powerful framework for anisotropic volume reconstruction in volume electron microscopy.
By recasting slice interpolation as dynamic 3D scene rendering with deformable Gaussian splatting, EMGauss removes the strong isotropy assumptions commonly embedded in previous approaches and provides a more principled, geometry-aware formulation of the reconstruction problem. 
Leveraging a self-supervised teacher–student bootstrapping scheme, our method learns directly from the target anisotropic slices without requiring external isotropic volumes or paired HR–LR data. This enables continuous, high-fidelity slice synthesis while suppressing artifacts that often arise in purely 2D formulations. Experiments demonstrate improved structural coherence across depth while preserving fine morphological details critical for vEM analysis. 
More broadly, EMGauss establishes a general and modality-agnostic paradigm for slice-to-volume reconstruction. The combination of deformable Gaussians, continuous depth modeling, and a self-supervised learning strategy offers a promising foundation for high-throughput 3D imaging in diverse scientific domains, enabling stable reconstruction without relying on large paired datasets or modality-specific supervision. We believe that the proposed framework opens up new possibilities for scalable, robust, and data-efficient 3D reconstruction pipelines across a wide range of planar acquisition systems.

% across a wide range of planar acquisition systems, including various microscopy and tomographic imaging settings.

\section*{Acknowledgements}
This work was supported by the National Natural Science Foundation of China (Grant 62250062), the Smart Grid National Science and Technology Major Project (Grant 2024ZD0801200), the Shanghai Municipal Science and Technology Major Project (Grant 2021SHZDZX0102), and the Fundamental Research Funds for the Central Universities.

% \clearpage
% \newpage
{
    \small
    \bibliographystyle{ieeenat_fullname}
    \bibliography{main}

@String(TOG= {ACM Trans. Graph.})

@String(AAAI = {AAAI})

@String(TOG   = {ACM TOG})

@article{kerbl20233d,
  title={3D Gaussian splatting for real-time radiance field rendering.},
  author={Kerbl, Bernhard and Kopanas, Georgios and Leimk{\"u}hler, Thomas and Drettakis, George},
  journal={ACM Trans. Graph.},
  volume={42},
  number={4},
  pages={139--1},
  year={2023}
}

@article{mildenhall2021nerf,
  title={Nerf: Representing scenes as neural radiance fields for view synthesis},
  author={Mildenhall, Ben and Srinivasan, Pratul P and Tancik, Matthew and Barron, Jonathan T and Ramamoorthi, Ravi and Ng, Ren},
  journal={Communications of the ACM},
  volume={65},
  number={1},
  pages={99--106},
  year={2021},
  publisher={ACM New York, NY, USA}
}

@inproceedings{wu20244d,
  title={4d gaussian splatting for real-time dynamic scene rendering},
  author={Wu, Guanjun and Yi, Taoran and Fang, Jiemin and Xie, Lingxi and Zhang, Xiaopeng and Wei, Wei and Liu, Wenyu and Tian, Qi and Wang, Xinggang},
  booktitle={Proceedings of the IEEE/CVF conference on computer vision and pattern recognition},
  pages={20310--20320},
  year={2024}
}

@inproceedings{yang2024deformable,
  title={Deformable 3d gaussians for high-fidelity monocular dynamic scene reconstruction},
  author={Yang, Ziyi and Gao, Xinyu and Zhou, Wen and Jiao, Shaohui and Zhang, Yuqing and Jin, Xiaogang},
  booktitle={Proceedings of the IEEE/CVF conference on computer vision and pattern recognition},
  pages={20331--20341},
  year={2024}
}

@article{muller2022instant,
  title={Instant neural graphics primitives with a multiresolution hash encoding},
  author={M{\"u}ller, Thomas and Evans, Alex and Schied, Christoph and Keller, Alexander},
  journal={ACM transactions on graphics (TOG)},
  volume={41},
  number={4},
  pages={1--15},
  year={2022},
  publisher={ACM New York, NY, USA}
}

@article{paszke2019pytorch,
  title={Pytorch: An imperative style, high-performance deep learning library},
  author={Paszke, Adam and Gross, Sam and Massa, Francisco and Lerer, Adam and Bradbury, James and Chanan, Gregory and Killeen, Trevor and Lin, Zeming and Gimelshein, Natalia and Antiga, Luca and others},
  journal={NeurIPS},
  volume={32},
  year={2019}
}

@article{kingma2014adam,
  title={Adam: A method for stochastic optimization},
  author={Kingma, Diederik P and Ba, Jimmy},
  journal={arXiv preprint arXiv:1412.6980},
  year={2014}
}

@inproceedings{heinrich2017deep,
  title={Deep learning for isotropic super-resolution from non-isotropic 3D electron microscopy},
  author={Heinrich, Larissa and Bogovic, John A and Saalfeld, Stephan},
  booktitle={International Conference on Medical Image Computing and Computer-Assisted Intervention},
  pages={135--143},
  year={2017},
  organization={Springer}
}

@article{collinson2023volume,
  title={Volume EM: a quiet revolution takes shape},
  author={Collinson, Lucy M and Bosch, Carles and Bullen, Anwen and Burden, Jemima J and Carzaniga, Raffaella and Cheng, Cheng and Darrow, Michele C and Fletcher, Georgina and Johnson, Errin and Narayan, Kedar and others},
  journal={Nature methods},
  volume={20},
  number={6},
  pages={777--782},
  year={2023},
  publisher={Nature Publishing Group US New York}
}

@article{glausier2025volume,
  title={Volume electron microscopy reveals 3D synaptic nanoarchitecture in postmortem human prefrontal cortex},
  author={Glausier, Jill R and Maier, Matthew and Bouchet-Marquis, Cedric and Wu, Ken and Banks-Tibbs, Tabitha and Melchitzky, Darlene and Ning, Jiying and Lewis, David A and Freyberg, Zachary},
  journal={iScience},
  volume={28},
  number={7},
  year={2025},
  publisher={Elsevier}
}

@article{li2025situ,
  title={In situ architecture of the intercellular organelle reservoir between epididymal epithelial cells by volume electron microscopy},
  author={Li, Xia and Qiao, Feng and Guo, Jiansheng and Jiang, Ting and Lou, Huifang and Li, Huixia and Xie, Gangcai and Wu, Hangjun and Wang, Weizhen and Pei, Ruoyu and others},
  journal={Nature Communications},
  volume={16},
  number={1},
  pages={1664},
  year={2025},
  publisher={Nature Publishing Group UK London}
}

@inproceedings{robert2025improving,
  title={Improving cell instance segmentation in scanning electron microscopy via semantic image synthesis},
  author={Robert, Florian and Calovoulos, Alexia and Facq, Laurent and Decoeur, Fanny and Gontier, Etienne and Grosset, Christophe F and de Senneville, Baudouin Denis},
  booktitle={2025 IEEE 22nd International Symposium on Biomedical Imaging (ISBI)},
  pages={1--5},
  year={2025},
  organization={IEEE}
}

@article{turegano2024tracing,
  title={Tracing nerve fibers with volume electron microscopy to quantitatively analyze brain connectivity},
  author={Turegano-Lopez, Marta and de Las Pozas, Felix and Santuy, Andrea and Rodriguez, Jose-Rodrigo and DeFelipe, Javier and Merchan-Perez, Angel},
  journal={Communications Biology},
  volume={7},
  number={1},
  pages={796},
  year={2024},
  publisher={Nature Publishing Group UK London}
}

@article{peddie2022volume,
  title={Volume electron microscopy},
  author={Peddie, Christopher J and Genoud, Christel and Kreshuk, Anna and Meechan, Kimberly and Micheva, Kristina D and Narayan, Kedar and Pape, Constantin and Parton, Robert G and Schieber, Nicole L and Schwab, Yannick and others},
  journal={Nature Reviews Methods Primers},
  volume={2},
  number={1},
  pages={51},
  year={2022},
  publisher={Nature Publishing Group UK London}
}

@inproceedings{cycleganir,
  title={Isotropic reconstruction of 3D EM images with unsupervised degradation learning},
  author={Deng, Shiyu and Fu, Xueyang and Xiong, Zhiwei and Chen, Chang and Liu, Dong and Chen, Xuejin and Ling, Qing and Wu, Feng},
  booktitle={International Conference on Medical Image Computing and Computer-Assisted Intervention},
  pages={163--173},
  year={2020},
  organization={Springer}
}

@inproceedings{pan2023diffuseir,
  title={DiffuseIR: diffusion models for isotropic reconstruction of 3D microscopic images},
  author={Pan, Mingjie and Gan, Yulu and Zhou, Fangxu and Liu, Jiaming and Zhang, Ying and Wang, Aimin and Zhang, Shanghang and Li, Dawei},
  booktitle={International Conference on Medical Image Computing and Computer-Assisted Intervention},
  pages={323--332},
  year={2023},
  organization={Springer}
}

@article{emdiffuse,
  title={Diffusion-based deep learning method for augmenting ultrastructural imaging and volume electron microscopy},
  author={Lu, Chixiang and Chen, Kai and Qiu, Heng and Chen, Xiaojun and Chen, Gu and Qi, Xiaojuan and Jiang, Haibo},
  journal={Nature Communications},
  volume={15},
  number={1},
  pages={4677},
  year={2024},
  publisher={Nature Publishing Group UK London}
}

@article{hagita2018super,
  title={Super-resolution for asymmetric resolution of FIB-SEM 3D imaging using AI with deep learning},
  author={Hagita, Katsumi and Higuchi, Takeshi and Jinnai, Hiroshi},
  journal={Scientific reports},
  volume={8},
  number={1},
  pages={5877},
  year={2018},
  publisher={Nature Publishing Group UK London}
}

@article{he2023isovem,
  title={IsoVEM: isotropic reconstruction for volume electron microscopy based on transformer},
  author={He, Jia and Zhang, Yan and Sun, Wenhao and Yang, Ge and Sun, Fei},
  journal={bioRxiv},
  pages={2023--11},
  year={2023},
  publisher={Cold Spring Harbor Laboratory}
}

@article{zhang2012noddi,
  title={NODDI: practical in vivo neurite orientation dispersion and density imaging of the human brain},
  author={Zhang, Hui and Schneider, Torben and Wheeler-Kingshott, Claudia A and Alexander, Daniel C},
  journal={Neuroimage},
  volume={61},
  number={4},
  pages={1000--1016},
  year={2012},
  publisher={Elsevier}
}

@article{beaulieu2002basis,
  title={The basis of anisotropic water diffusion in the nervous system--a technical review},
  author={Beaulieu, Christian},
  journal={NMR in Biomedicine: An International Journal Devoted to the Development and Application of Magnetic Resonance In Vivo},
  volume={15},
  number={7-8},
  pages={435--455},
  year={2002},
  publisher={Wiley Online Library}
}

@article{weigert2018content,
  title={Content-aware image restoration: pushing the limits of fluorescence microscopy},
  author={Weigert, Martin and Schmidt, Uwe and Boothe, Tobias and M{\"u}ller, Andreas and Dibrov, Alexandr and Jain, Akanksha and Wilhelm, Benjamin and Schmidt, Deborah and Broaddus, Coleman and Culley, Si{\^a}n and others},
  journal={Nature methods},
  volume={15},
  number={12},
  pages={1090--1097},
  year={2018},
  publisher={Nature Publishing Group US New York}
}

@inproceedings{krull2019noise2void,
  title={Noise2void-learning denoising from single noisy images},
  author={Krull, Alexander and Buchholz, Tim-Oliver and Jug, Florian},
  booktitle={Proceedings of the IEEE/CVF conference on computer vision and pattern recognition},
  pages={2129--2137},
  year={2019}
}

@article{zhao2024application,
  title={The application and development of electron microscopy for three-dimensional reconstruction in life science: a review},
  author={Zhao, Jingjing and Yu, Xiaoping and Shentu, Xuping and Li, Danting},
  journal={Cell and Tissue Research},
  volume={396},
  number={1},
  pages={1--18},
  year={2024},
  publisher={Springer}
}

@article{zhang20242d,
  title={2D Gaussian Splatting for Image Compression},
  author={Zhang, Pingping and Liu, Xiangrui and Wang, Meng and Wang, Shiqi and Kwong, Sam and others},
  journal={APSIPA Transactions on Signal and Information Processing},
  volume={13},
  number={6},
  year={2024},
  publisher={Now Publishers, Inc.}
}

@inproceedings{zhang2024gaussianimage,
  title={Gaussianimage: 1000 fps image representation and compression by 2d gaussian splatting},
  author={Zhang, Xinjie and Ge, Xingtong and Xu, Tongda and He, Dailan and Wang, Yan and Qin, Hongwei and Lu, Guo and Geng, Jing and Zhang, Jun},
  booktitle={European Conference on Computer Vision},
  pages={327--345},
  year={2024},
  organization={Springer}
}

@inproceedings{hu2025gaussiansr,
  title={Gaussiansr: High fidelity 2d gaussian splatting for arbitrary-scale image super-resolution},
  author={Hu, Jintong and Xia, Bin and Chen, Bin and Yang, Wenming and Zhang, Lei},
  booktitle={Proceedings of the AAAI Conference on Artificial Intelligence},
  volume={39},
  number={4},
  pages={3554--3562},
  year={2025}
}

@inproceedings{zhu2025large,
  title={Large Images Are Gaussians: High-Quality Large Image Representation with Levels of 2D Gaussian Splatting},
  author={Zhu, Lingting and Lin, Guying and Chen, Jinnan and Zhang, Xinjie and Jin, Zhenchao and Wang, Zhao and Yu, Lequan},
  booktitle={Proceedings of the AAAI Conference on Artificial Intelligence},
  volume={39},
  number={10},
  pages={10977--10985},
  year={2025}
}

@inproceedings{zhu2017unpaired,
  title={Unpaired image-to-image translation using cycle-consistent adversarial networks},
  author={Zhu, Jun-Yan and Park, Taesung and Isola, Phillip and Efros, Alexei A},
  booktitle={Proceedings of the IEEE international conference on computer vision},
  pages={2223--2232},
  year={2017}
}

@article{takemura2015synaptic,
  title={Synaptic circuits and their variations within different columns in the visual system of Drosophila},
  author={Takemura, Shin-ya and Xu, C Shan and Lu, Zhiyuan and Rivlin, Patricia K and Parag, Toufiq and Olbris, Donald J and Plaza, Stephen and Zhao, Ting and Katz, William T and Umayam, Lowell and others},
  journal={Proceedings of the National Academy of Sciences},
  volume={112},
  number={44},
  pages={13711--13716},
  year={2015},
  publisher={National Academy of Sciences}
}

@inproceedings{lucchi2013learning,
  title={Learning for structured prediction using approximate subgradient descent with working sets},
  author={Lucchi, Aur{\'e}lien and Li, Yunpeng and Fua, Pascal},
  booktitle={Proceedings of the IEEE Conference on Computer Vision and Pattern Recognition},
  pages={1987--1994},
  year={2013}
}

@article{phelps2021reconstruction,
  title={Reconstruction of motor control circuits in adult Drosophila using automated transmission electron microscopy},
  author={Phelps, Jasper S and Hildebrand, David Grant Colburn and Graham, Brett J and Kuan, Aaron T and Thomas, Logan A and Nguyen, Tri M and Buhmann, Julia and Azevedo, Anthony W and Sustar, Anne and Agrawal, Sweta and others},
  journal={Cell},
  volume={184},
  number={3},
  pages={759--774},
  year={2021},
  publisher={Elsevier}
}

@article{wang2004image,
  title={Image quality assessment: from error visibility to structural similarity},
  author={Wang, Zhou and Bovik, Alan C and Sheikh, Hamid R and Simoncelli, Eero P},
  journal={IEEE transactions on image processing},
  volume={13},
  number={4},
  pages={600--612},
  year={2004},
  publisher={IEEE}
}

@article{zhang2011fsim,
  title={FSIM: A feature similarity index for image quality assessment},
  author={Zhang, Lin and Zhang, Lei and Mou, Xuanqin and Zhang, David},
  journal={IEEE transactions on Image Processing},
  volume={20},
  number={8},
  pages={2378--2386},
  year={2011},
  publisher={IEEE}
}

@inproceedings{carata2011improving,
  title={Improving the visualization of electron-microscopy data through optical flow interpolation},
  author={Carata, Lucian and Shao, Dan and Hadwiger, Markus and Groeller, Eduard},
  booktitle={Proceedings of the 27th spring conference on computer graphics},
  pages={103--110},
  year={2011}
}

@article{gonzalez2022optical,
  title={Optical flow driven interpolation for isotropic FIB-SEM reconstructions},
  author={Gonz{\'a}lez-Ruiz, Vicente and Garc{\'\i}a-Ortiz, Juan Pablo and Fern{\'a}ndez-Fern{\'a}ndez, Mar{\'\i}a Rosario and Fern{\'a}ndez, Jos{\'e}-Jes{\'u}s},
  journal={Computer Methods and Programs in Biomedicine},
  volume={221},
  pages={106856},
  year={2022},
  publisher={Elsevier}
}

@inproceedings{ferede2025z,
  title={Z-upscaling: Optical Flow Guided Frame Interpolation for Isotropic Reconstruction of 3D EM Volumes},
  author={Ferede, Fisseha A and Khalighifar, Ali and John, Jaison and Venkataraman, Krishnan and Khairy, Khaled},
  booktitle={2025 IEEE 22nd International Symposium on Biomedical Imaging (ISBI)},
  pages={1--5},
  year={2025},
  organization={IEEE}
}

@article{joshi2025interpolai,
  title={InterpolAI: deep learning-based optical flow interpolation and restoration of biomedical images for improved 3D tissue mapping},
  author={Joshi, Saurabh and Forjaz, Andr{\'e} and Han, Kyu Sang and Shen, Yu and Queiroga, Vasco and Selaru, Florin A and G{\'e}rard, Marie and Xenes, Daniel and Matelsky, Jordan and Wester, Brock and others},
  journal={Nature Methods},
  pages={1--12},
  year={2025},
  publisher={Nature Publishing Group US New York}
}

@article{briggman2006towards,
  title={Towards neural circuit reconstruction with volume electron microscopy techniques},
  author={Briggman, Kevin L and Denk, Winfried},
  journal={Current opinion in neurobiology},
  volume={16},
  number={5},
  pages={562--570},
  year={2006},
  publisher={Elsevier}
}

@article{hayworth2014imaging,
  title={Imaging ATUM ultrathin section libraries with WaferMapper: a multi-scale approach to EM reconstruction of neural circuits},
  author={Hayworth, Kenneth J and Morgan, Josh L and Schalek, Richard and Berger, Daniel R and Hildebrand, David GC and Lichtman, Jeff W},
  journal={Frontiers in neural circuits},
  volume={8},
  pages={68},
  year={2014},
  publisher={Frontiers Media SA}
}

@article{denk2004serial,
  title={Serial block-face scanning electron microscopy to reconstruct three-dimensional tissue nanostructure},
  author={Denk, Winfried and Horstmann, Heinz},
  journal={PLoS biology},
  volume={2},
  number={11},
  pages={e329},
  year={2004},
  publisher={Public Library of Science San Francisco, USA}
}

@article{bushby2011imaging,
  title={Imaging three-dimensional tissue architectures by focused ion beam scanning electron microscopy},
  author={Bushby, Andrew J and P'ng, Kenneth MY and Young, Robert D and Pinali, Christian and Knupp, Carlo and Quantock, Andrew J},
  journal={Nature protocols},
  volume={6},
  number={6},
  pages={845--858},
  year={2011},
  publisher={Nature Publishing Group UK London}
}

@article{ravi2024sam2,
  title={SAM 2: Segment Anything in Images and Videos},
  author={Ravi, Nikhila and Gabeur, Valentin and Hu, Yuan-Ting and Hu, Ronghang and Ryali, Chaitanya and Ma, Tengyu and Khedr, Haitham and R{\"a}dle, Roman and Rolland, Chloe and Gustafson, Laura and Mintun, Eric and Pan, Junting and Alwala, Kalyan Vasudev and Carion, Nicolas and Wu, Chao-Yuan and Girshick, Ross and Doll{\'a}r, Piotr and Feichtenhofer, Christoph},
  journal={arXiv preprint arXiv:2408.00714},
  url={https://arxiv.org/abs/2408.00714},
  year={2024}
}
}
%%%%%%%%%%%%%%%%%%%%%%%%%%%%%% main paper %%%%%%%%%%%%%%%%%%%%%%%%%%%%%%

%%%%%%%%%%%%%%%%%%%%%%%%%%%%%% supp %%%%%%%%%%%%%%%%%%%%%%%%%%%%%%
% \input{sec/X_supp}
% \clearpage
% \newpage
% {
%     \small
%     \bibliographystyle{ieeenat_fullname}
%     \bibliography{main}
% }
%%%%%%%%%%%%%%%%%%%%%%%%%%%%%% supp %%%%%%%%%%%%%%%%%%%%%%%%%%%%%%

\end{document}